\documentclass{article} 
\usepackage{iclr2023_conference,times}

\usepackage{amsmath,amsfonts,bm}

\def\tabref#1{table~\ref{#1}}
\def\Tabref#1{Table~\ref{#1}}
\def\figref#1{figure~\ref{#1}}

\def\appref#1{Appendix~\ref{#1}}
\def\secref#1{section~\ref{#1}}

\def\eqref#1{equation~\ref{#1}}

\def\1{\bm{1}}

\DeclareMathAlphabet{\mathsfit}{\encodingdefault}{\sfdefault}{m}{sl}
\SetMathAlphabet{\mathsfit}{bold}{\encodingdefault}{\sfdefault}{bx}{n}

\newcommand{\normltwo}{L^2}

\newcommand{\range}{\text{\textasciitilde}}
\newcommand{\eg}{{\it e.g.}}

\def\decoder{{\mathcal{B}}}

\renewcommand{\paragraph}[1]{\textbf{#1}~\,}

\newcommand{\first}{{(i) }}
\newcommand{\second}{{(ii) }}
\newcommand{\third}{{(iii) }}
\newcommand{\fourth}{{(iv) }}

\usepackage[utf8]{inputenc} 
\usepackage[T1]{fontenc}    
\definecolor{citecolor}{rgb}{0.052,0.11,0.508}
\definecolor{linkcolor}{rgb}{0.9, 0.06, 0.06}
\usepackage[colorlinks=true,citecolor=citecolor, linkcolor=linkcolor]{hyperref}
\usepackage{url}            
\usepackage{booktabs}       
\usepackage{amsfonts}       
\usepackage{nicefrac}       
\usepackage{microtype}      

\usepackage{algorithm}
\usepackage{algorithmic}
\usepackage{xcolor}
\usepackage{color}
\usepackage{colortbl}
\usepackage{subfigure}
\usepackage{multirow}
\usepackage{bbm}
\usepackage{caption}
\usepackage{mathtools}
\usepackage{wrapfig}

\usepackage{listings}
\usepackage{color}

\definecolor{dkgreen}{rgb}{0,0.6,0}
\definecolor{gray}{rgb}{0.5,0.5,0.5}
\definecolor{mauve}{rgb}{0.58,0,0.82}

\definecolor{deepblue}{rgb}{0,0,0.5}
\definecolor{deepred}{rgb}{0.6,0,0}
\definecolor{deepgreen}{rgb}{0,0.5,0}

\DeclareFixedFont{\ttb}{T1}{txtt}{bx}{n}{9} 
\DeclareFixedFont{\ttm}{T1}{txtt}{m}{n}{9}  

\newcommand\pythonstyle{\lstset{frame=tb,
  language=Python,
  basicstyle={\small\ttm},
  morekeywords={self},              
  keywordstyle=\ttb\color{deepblue},
  emph={MyClass,__init__},          
  emphstyle=\ttb\color{deepred},    
  stringstyle=\color{deepgreen},
  commentstyle=\color{dkcyan},
  numberstyle=\tiny\color{gray},
  aboveskip=3mm,
  belowskip=3mm,
  showstringspaces=false,
  frame=tb,                         
  columns=flexible,
  numbers=none,
  breaklines=true,
  breakatwhitespace=true,
  tabsize=3
}}

\lstnewenvironment{python}[1][]
{
\pythonstyle
\lstset{#1}
}
{}

\usepackage{colortbl}
\usepackage{tablefootnote}

\definecolor{blgrey}{rgb}{0.6,0.6,0.6}
\definecolor{bblue}{rgb}{0.855,0.933,0.98}
\definecolor{dblue}{HTML}{5297D6}
\definecolor{gainred}{rgb}{0.1,0.5,0.3}
\definecolor{hblue}{HTML}{4292C1}

\definecolor{convcolor}{HTML}{4E8AC6}
\definecolor{rescolor}{HTML}{8DA0CB}
\definecolor{vitcolor}{HTML}{7A3FAF}

\newcommand{\bluebg}[1]{\colorbox{bblue!88}{#1}}
\newcommand{\grayfg}[1]{\textcolor{gray!60}{#1}}
\newcommand{\ablanum}[1]{\textcolor{dblue}{#1}}
\newcommand{\ablaref}[1]{row \textcolor{dblue}{#1}}

\usepackage{amssymb}
\usepackage{pifont}
\usepackage{bbding}
\newcommand{\ncr}{\ding{55}}
\renewcommand{\ncr}{}
\newcommand{\nck}{\ding{51}}

\newcommand{\tabbl}{\midrule \rowcolor{lightgray!17}}

\usepackage{enumitem}

\def\smallcaption{\small}

\newcommand{\rbt}[1]{{#1}}

\title{
\mbox{Designing BERT for Convolutional Networks:}\\Sparse and Hierarchical Masked Modeling
}

\author{Keyu Tian$^{1,2}$, \,\,~~\qquad\qquad Yi Jiang$^{2}$,   \,\,~~~\quad\qquad\qquad Qishuai Diao$^{2}$, \\
\textbf{Chen Lin$^{3}$,}    ~~~~\quad\qquad\qquad \textbf{Liwei Wang$^{1}$,} ~~\qquad\qquad \textbf{Zehuan Yuan$^{2}$} \\
$^1$Peking University \quad\qquad $^2$Bytedance Inc. ~~~\quad\qquad $^3$University of Oxford \\
\texttt{\footnotesize keyutian@stu.pku.edu.cn, \{jiangyi.enjoy,diaoqishuai\}@bytedance.com,} \\
\texttt{\footnotesize chen.lin@eng.ox.ac.uk, wanglw@pku.edu.cn, yuanzehuan@bytedance.com} \\
}

\iclrfinalcopy 
\begin{document}

\maketitle
\vspace{-4pt}

\vspace{-8pt}
\def\absemph{\underline}

\begin{abstract}
\vspace{-1pt}
We identify and overcome two key obstacles in extending the success of BERT-style pre-training, or masked image modeling, to convolutional networks (convnets):
(i) convolution operation cannot handle irregular, randomly masked input images;
(ii) the single-scale nature of BERT pre-training is inconsistent with convnet's hierarchical structure.
For (i), we treat unmasked pixels as sparse voxels of 3D point clouds and use sparse convolution to encode.
This is the first use of sparse convolution for 2D masked modeling.
For (ii), we develop a hierarchical decoder to reconstruct images from multi-scale encoded features.
Our method, called \textit{Spar}se mas\textit{K}ed modeling (\textit{SparK}), is general: it can be used directly on any convolutional model without backbone modifications.
We validate it on both classical (ResNet) and modern (ConvNeXt) models:
on three downstream tasks, it surpasses both state-of-the-art contrastive learning and transformer-based masked modeling by similarly large margins (around $+1.0\%$).
The improvements on object detection and instance segmentation are more significant (up to $+3.5\%$), validating the strong transferability of features learned.
We also find its favorable scaling behavior by observing more gains on larger networks.
All this evidence reveals a promising future of generative pre-training on convnets.
Codes and models are released at {\small \url{https://github.com/keyu-tian/SparK}}.
\vspace{-1pt}
\end{abstract}

\vspace{-8pt}

\section{Introduction} \label{sec:intro}
The pretrain-finetune paradigm in natural language processing (NLP), popularized by BERT \citep{bert,unilm,electra} and GPT \citep{gpt1,gpt2}, is remarkably effective and thus long envied by our vision community.
It is the emerging masked image modeling \citep{beit,mae,simmim,cae} initially extends the success of BERT \textit{from language transformers to vision transformers} (ViTs).
\mbox{A bold move that} increases the mask ratio to a staggering level (60\range 75\%) is largely credited with this success \citep{mae,simmim}.
And the field of visual self-supervised learning on ViTs \citep{vit,swin} has now shifted from contrastive learning \citep{byol,mocov3,dino} to BERT-style masked modeling or a fusion of them \citep{ibot}.

Nonetheless, extending the success of BERT pre-training\,~\textit{from transformers to convolutional networks} (convnets) is a wonderful, but unrealized vision.
\rbt{The pioneering work \citep{inpainting1,inpainting2} preceded BERT but performed much worse than supervised pre-training.
There have been efforts over the past year trying to port BERT to convnets, yet eventually compromise on proposing a non-convolutional model \citep{convmae} or non-masked modeling \citep{cim}.}
One may therefore ask: \textit{what exactly is preventing the application of BERT to convnets?}

\begin{figure}[t]
\begin{center}
   \includegraphics[width=0.855\linewidth]{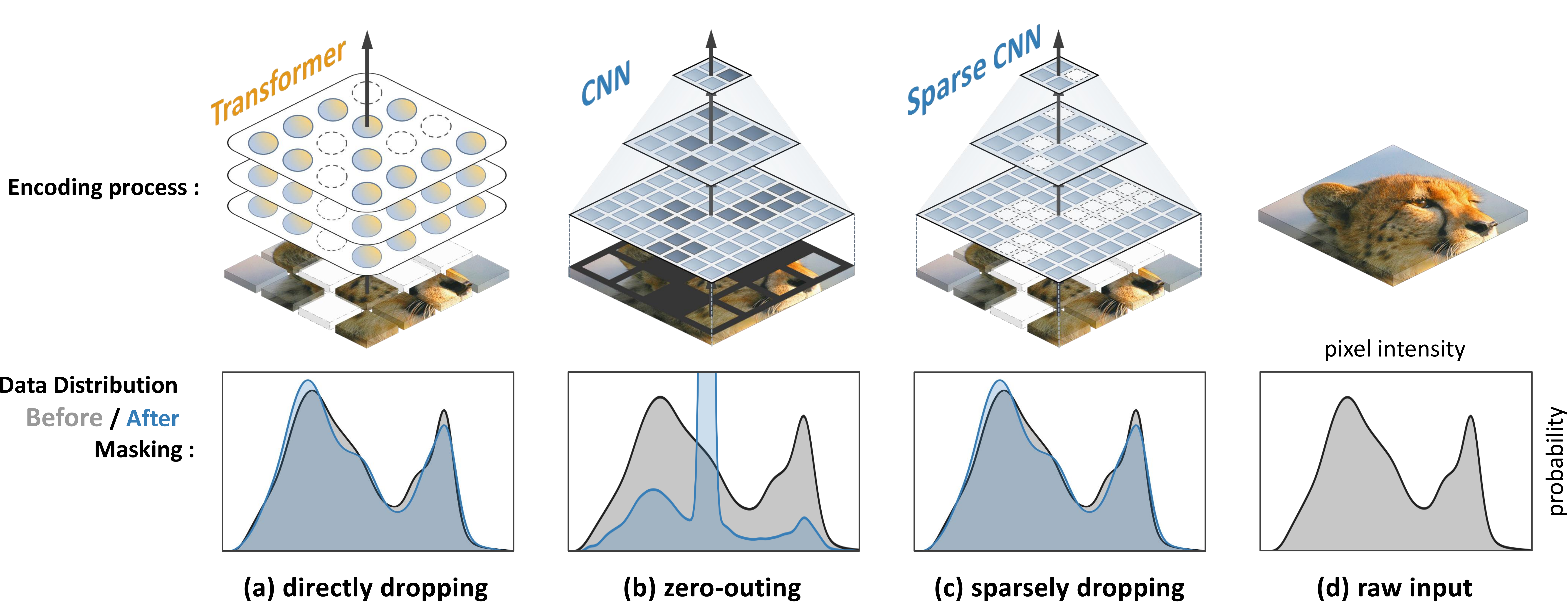}
\end{center}
\vspace{-5pt}
\caption{\smallcaption
\textbf{Different masking strategies} with pixel intensity histograms plotted before (in gray) and after (blue) masking.
(b) is a straightforward idea to apply masked modeling to convnets, which results in a distribution shift.
(a) illustrates MAE \citep{mae} that has no such side effect thanks to the transformer's ability to process variable-length input.
We propose (c) to adapt convnets to irregular masked input without a distribution shift.
\vspace{-8pt}
}
\label{fig:intro}
\end{figure}

We try to conclude that in essence, the difficulty is rooted in the \textit{gap} between language and vision in terms of data processing \citep{bateman2014text,mae}.
Typical NLP models like recurrent networks or transformers process text as a variable-length sequence of words (well-defined semantic units),
while convnets have to recognize objects of different sizes from raw pixels (like ``units'' at different scales).
This huge disparity rises two challenges:
\textbf{\first Removing the information of masked ``words'' is difficult for convnets.}
For ViTs, an input image is divided into several non-overlapping patches.
Simply dropping masked patches or replacing them with mask tokens can remove the information.
This ease relies on transformer being able to handle \textit{irregular (variable-length)} and \textit{non-overlapping} patches, thus cannot be achieved on convnets as they not only operate on \textit{regular} grids, but also perform sliding window \textit{with overlapping}.
One may zero-out all masked pixels and feed this ``mosaic'' into a convnet.
This, however, leads to a severe data distribution shift (as in \figref{fig:intro}) and other issues (discussed later in \secref{sec:method:mask} and \figref{fig:ssc}), thus cannot be an ideal solution.
\textbf{\second Single-scale algorithm cannot learn multi-scale (hierarchical) features.}
Multi-scale structure has long been a gold standard in computer vision, which allows visual processing systems like SIFT descriptors \citep{SIFT,surf} and pyramid networks \citep{spp,FPN} to cope with changes in object scale.
In contrast, masked modeling from NLP originally works in a single-scale manner. 
Applying it directly on convnets will miss the advantage of model hierarchy.

In this work, we clear the hurdles above and make BERT suitable for convnet by proposing \textbf{Spar}se mas\textbf{K}ed modeling with hierarchy (SparK).
It first randomly masks images in a patch-wise manner.
Observing the sparse nature of point clouds coincides with these unmasked patches, we treat them as a flatten 3D point cloud and use sparse convolution for encoding.
This strategy accurately eliminates the information of masked parts whilst allowing convnets to easily handle irregular masked images.
For decoding, we fill in all empty positions on the mult-scale features with mask embeddings, and feed them into a multi-scale decoder.
This way, we can embrace the advantage of convnet's hierarchy.

SparK is a general method that does not limit the specific encoder to be pre-trained.
We test it with two representative convnet famlies: classical ResNets \citep{resnet} and modern ConvNeXts \citep{convnext}.
All models benefit from SparK, with more gains on larger models that demonstrates its favorable scaling ability.
On standard downstream tasks (classification, object detection and instance segmentation), convnet-based SparK outperforms both \first state-of-the-art contrastive learning and \second transformer-based masked modeling by similarly large margins (around $+1.0\%$).
The improvements over COCO baselines are more significant than those on ImageNet (up to $+3.5\%$), indicating the representations learned by SparK are highly transferable.
\vspace{-2pt}
To summarize, SparK provides:
\begin{itemize}[leftmargin=20pt,topsep=4pt,itemsep=2pt]
    \item The first BERT-style pre-training method that can be used directly on any convnets without backbone modifications, overcoming their inability to handle irregular masked inputs.
    \item The insights into designing generative pre-training for convnets, \eg, the first use of sparse convolution for masked image modeling, and the hierarchical design for BERT-style pre-training.
    \item A leap in convnet's performance across downstream tasks (up to $+3.5$ points), showing the promise of extending the success of transformer's pretrain-finetune paradigm to convnets.
\end{itemize}
\vspace{-2pt}

The computer vision community has recently paid more attention to vision transformers, while convnets no longer appear in the spotlight \citep{swin,mae}.
Nonetheless, these neural networks that condense the essence (scale- and translation-equivariant; locality; weight-sharing; hardware-friendly) of many classical vision processing systems \citep{SIFT,bow,farabet2010hardware} still play an irreplaceable role in tackling a wide range of real-world tasks that are more challenging or structural than classification \citep{stn,structural2,convnext}.
We hope SparK's inspiring performance will prompt us to revisit convnets as generic backbones for our community, and inspire more future arts in exploiting their potential via generative pre-training.

\section{Related Work} \label{sec:related}
\subsection{Hierarchical visual processing systems}
\paragraph{Hierarchical structure} is acknowledged as a gold standard for visual representation systems.
Many fundamental handcrafted feature descriptors \citep{SIFT,surf,orb} extract multi-scale visual representations via scale-space extremum on feature pyramid (say, octave).
The crux behind this hierarchical design is to extract scale-invariant (or equivariant) features, thus, allows the system to cope with varying object sizes (scales).
Widely used in visual tasks \citep{DPM,sparsecoding}, these descriptors also motivate the design principles of convolutional networks \citep{resnet,efficientnet,convnext}.
Some recent arts also elaborately design hierarchical modules that allow the information aggregation at different granularities to better tackle detection and segmentation tasks using convnets \citep{FCN,ssd,FPN}.

\subsection{Recent progress on visual self-supervised learning}

\paragraph{Recently, the contrastive learning} formulates self-supervise learning as an instance classification task \citep{cpc,moco,simclr}.
Efforts have been made \citep{byol,swav,simsiam} to overcome the core issue of mode collapse.
More advanced methods are developed since then \citep{infomin,barlow,mocov3,dino}, and this line of work had dominated the area of visual unsupervised learning until masked generative pre-training along with the vision transformer architecture came into view.

\paragraph{Masked image modeling,} inspired by the recent success of masked language modeling in natural language processing (NLP) \citep{bert,roberta}, has attracted growing interest for visual pre-training.
The pioneering work (\cite{beit}) pre-trains vision transformers by learning to predict token indices of masked patches.
\cite{mae} ingeniously takes advantage of transformer's ability to handle variable-length inputs and implements an efficient and scalable method.
Both \cite{mae} and \cite{simmim} regress raw RGBs to simplify the pre-training, while \cite{maskfeat} selects HOG \citep{HOG} as targets due to their rich semantics.
\cite{convmae} designs a transformer with a heavier patchifier to perform masked modeling.
So far, enormous studies have successfully verified the efficacy of these algorithms on vision transformers \citep{ibot,cae,cim}.
However, on the other hand, their methodology is almost the same as that in NLP \citep{bert,roberta}, and is therefore difficult to be used for hierarchical convolutional models -- on convnets, contrastive learning still remains state-of-the-art.

\subsection{Sparse convolution for visual representation}

Convolution is widely used in 2D computer vision \citep{HOG,resnet}, which typically performs sliding window on regular grids (pixels).
When facing with 3D point clouds, this operator quickly becomes unaffordable due to the cubic increasing number of grids (voxels).
Considering point clouds are highly sparse and irregular, one can skip all empty voxels for speed.
This motivates the \textbf{sparse convolution} (sparseconv) \citep{spcn}, which is heavily used in modern convnets for 3D visual tasks \citep{sp3d1,sp3d2}.
\rbt{Minkowski Engine \citep{Minkowski} is one of the most common sparseconv frameworks.
Some prior arts \citep{r3_cite} also tried to introduce sparseconv for faster 2D visual understanding.
And in this work, we have observed the similarity between 3D point clouds and 2D masked images in BERT-style pre-training. We thus use sparseconv, for the first time, with the purpose of ``facilitating the adaptation of convnet to BERT masked modeling'', rather than of ``speeding up the computation of convolution''.}

\section{Approach} \label{sec:method}

\begin{figure}[ht]
\begin{center}
   \includegraphics[width=0.84\linewidth]{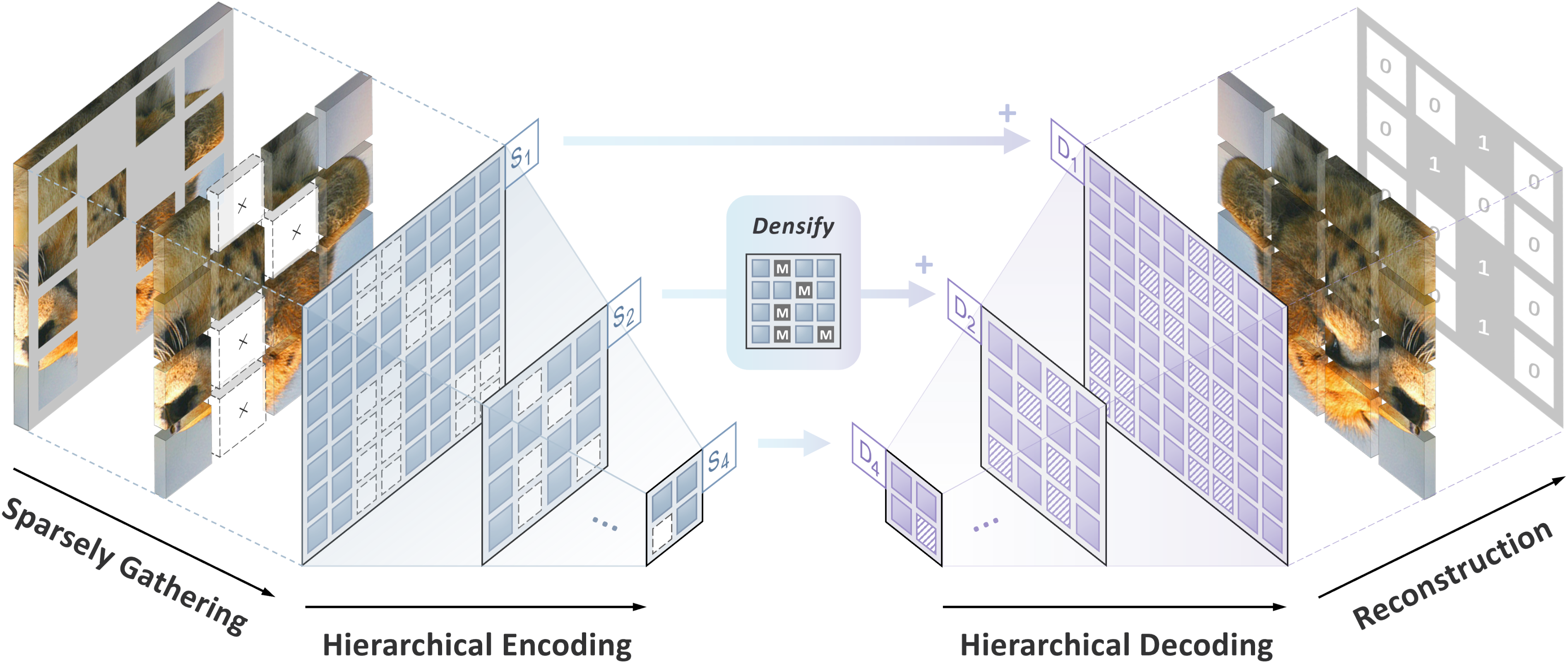}
\end{center}
\vspace{-5pt}
\caption{\smallcaption
\textbf{Sparse masked modeling with hierarchy.}\,
To adapt convolution to irregular masked input, visible patches are gathered into a sparse image and encoded by sparse convolution.
To pre-train a hierarchical encoder, we employ a UNet-style architecture to decode multi-scale sparse feature maps, where all empty positions are filled with mask embedding.
This ``densifying'' is necessary to reconstruct a dense image.
Only the regression loss on masked patches will be optimized.
After pre-training, only the encoder is used for downstream tasks.
\vspace{-6pt}
}
\label{fig:main}
\end{figure}

Illustrated in \figref{fig:main}, our SparK framework aims to pre-train a convolutional network encoder via hierarchical masked image modeling -- masking a portion of image and learning to recover it.
We are going to detail SparK by introducing a sparse masking strategy (\secref{sec:method:mask}), a hierarchical encoder-decoder architecture (\secref{sec:method:enc}), and the optimization target of SparK pre-training (\secref{sec:method:loss}).

\vspace{-4pt}

\subsection{Sparsely Gathering Unmasked Patches} \label{sec:method:mask}

We start by the patch-wise masking strategy widely used in masked image modeling.
An image is divided into several non-overlapping square patches, each of which will then be masked independently with a given probability called mask ratio.
The key to a masked image modeling algorithm is how to eliminate the pixel information from these masked patches.

\begin{figure}[t]
\begin{center}
   \includegraphics[width=0.84\linewidth]{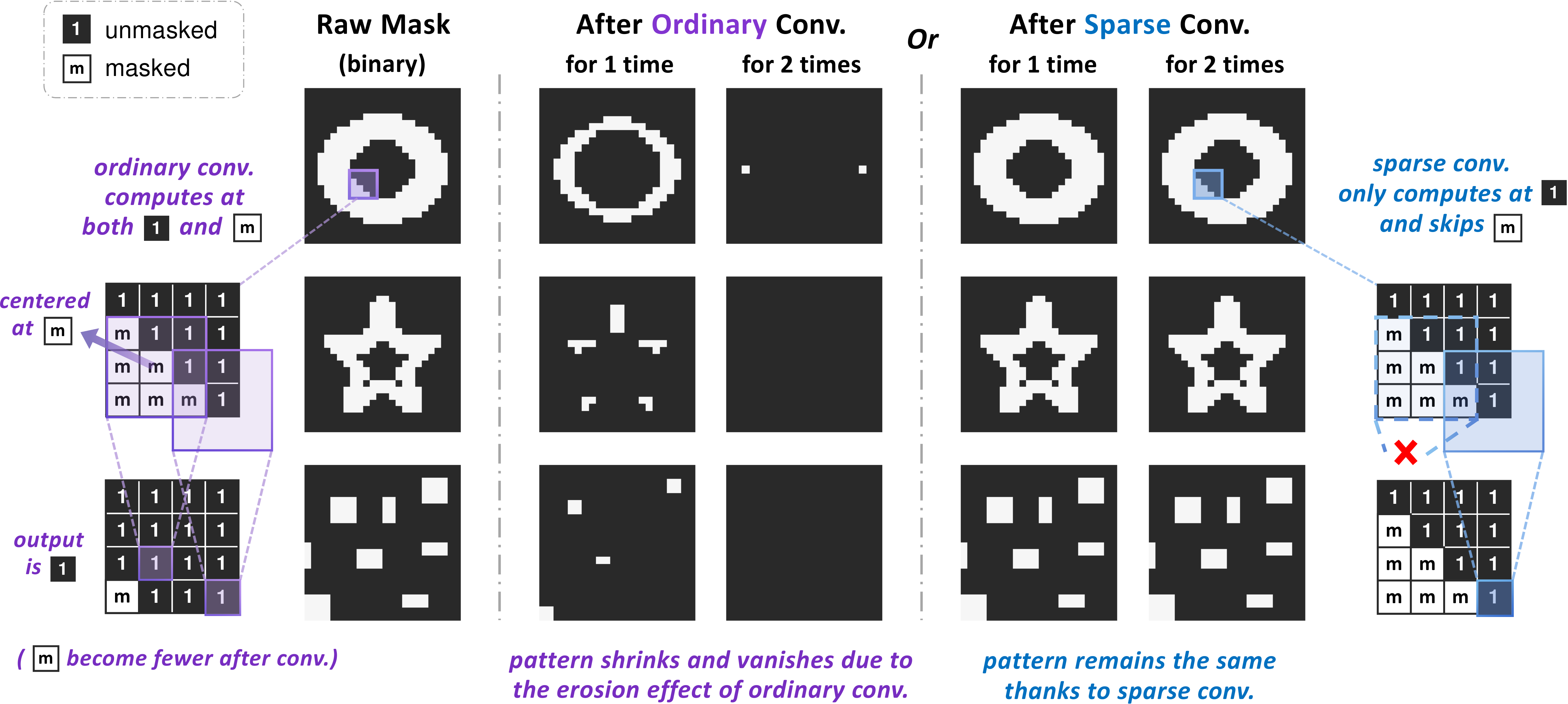}
\end{center}
\vspace{-5pt}
\caption{\smallcaption
\textbf{Using sparse convolution to address ``mask pattern vanishing'' issue.}
Three mask examples are shown.
As in left, when computing ordinary ``dense'' convolution centered at a zero (masked) position, the result would be non-zero if the filter covers any non-zero (unmasked) points.
Repeating this convolution will \textit{erode} masked regions (zero positions) and \textit{dilate} the unmasked ones, leading to the vanishing problem. 
\rbt{We use sparse convolution to overcome this undesired property by skipping all masked positions and keeping the mask pattern.}
\vspace{-5pt}
}
\label{fig:ssc}
\end{figure}

\paragraph{Previous transformer-based masked modeling} can easily eliminate the information by directly removing masked patches or replacing them with a mask token.
This ease relies on the fact that vision transformers are born to handle \textit{irregular} (variable-length) input and operate on \textit{non-overlapping} image patches.
Since convnets cannot do this, new approaches have to be sought.
A straightforward idea is to set all masked pixels to zero and feed this image to a convnet.
This, however, has three evident shortcomings:
\first the computation on masked regions is redundant;
\second it would disturb the data distribution of pixel values, as illustrated in \figref{fig:intro};
\third the patterns on mask maps will vanish after applying several convolutions to this zero-out masked image.
We examine problem \third in \figref{fig:ssc}, where we also give our solution.
Note that this problem is particularly acute when using modern deep convnets due to the large number of successive convolutional blocks.

\paragraph{To overcome the problems, we propose to sparsely gather all unmasked patches} into a sparse image, and then use \textbf{sparse convolutions}\footnote{By ``sparse convolution'', we mean the \textit{submanifold sparse convolution} that computes only when the kernel center covers a non-empty element. Please refer to \cite{ssc} for more details.} to encode it.
This strategy:
\first ensures no information is leaked;\,
\second can be applied directly to any convnet without backbone modifications;\,
\third is efficient as sparse convolution computes only at visible places;
\fourth solves the aforementioned issues of ``pixel distribution shift'' and ``mask pattern vanishing''. 
As shown in \figref{fig:ssc}, sparse convolution will skip all masked positions on sparse feature maps, and only computes at unmasked points.
This helps to prevent the shape of the mask pattern from changing with convolution, thus ensures a consistent masking effect and ratio throughout all convolution layers.
Another fact is that when fine-tuning, all sparse convolutional layers can be naturally reduced to ordinary dense ones.
This is true because dense images are actually the special cases of sparse images that have no ``holes''.

\subsection{Hierarchical Encoding and Decoding} \label{sec:method:enc}

\paragraph{By ``hierarchical'' encoding,} we mean the encoder will generate a set of feature maps with different resolutions, namely different \textit{scales}. 
Taking a ResNet-style model for example, it typically contains 4 stages each with a series of convolutional blocks and a downsampling module.
The feature resolution is downsampled by a factor of 2 after every stage.
For an image shaped as $H\times W$, a ResNet-50 produces feature maps at 4 scales with resolutions of $\frac{H}{4}\times \frac{W}{4}$, $\frac{H}{8}\times \frac{W}{8}$, $\frac{H}{16}\times \frac{W}{16}$, and $\frac{H}{32}\times \frac{W}{32}$.
Let $S_1$, $S_2$, $S_3$, and $S_4$ be these sparse features, respectively.
They will be used to decode.

\paragraph{Overall, the decoder} follows the design of UNet \citep{unet}.
We use a relatively light decoder that contains three successive blocks $\{\decoder_3, \decoder_2, \decoder_1\}$ with upsampling layers.
Before reconstructing a dense image, it is necessary to \textit{fill in} all the empty positions on sparse feature maps.
This is called ``\textbf{densifying}''.
Taking the smallest sparse feature $S_4$ as example, all empty positions (inactive sites) on $S_4$ are filled with a mask embedding $\texttt{[}\texttt{M}_4\texttt{]}$ to get a dense feature $S^\prime_4$.
A projection layer $\phi_4$ is applied then, in case encoder and decoder have different network widths:
\begin{align}
    D_4 = \phi_4(S^\prime_4).
\end{align}
So $D_4$ is the input of decoder's first block $\decoder_3$. It has the same resolution of $S_4$ with $\frac{H}{32}\times \frac{W}{32}$.
Similarly, we can get $D_3$, $D_2$, and $D_1$ (with shapes of $\frac{H}{16}\times \frac{W}{16}$, $\frac{H}{8}\times \frac{W}{8}$, $\frac{H}{4}\times \frac{W}{4}$) via: 
\begin{align}
    D_i = \decoder_i(D_{i+1}) \,+\, \phi_i(S^\prime_i) \quad\quad (\forall\, i \in \{3,2,1\} ) .
\end{align}
Note that four different mask embeddings $\texttt{[}\texttt{M}_{4\range 1}\texttt{]}$ and projection layers $\phi_{4\range 1}$ are required: they belong to different scales, and may have different network widths.
The final output of decoder is $D_1$.

\subsection{Optimization Target and Transferring to Downstream} \label{sec:method:loss}

\paragraph{To reconstruct an image from $D_1$,} a head module $h$ is needed, which should include two more upsampling layers to reach the original resolution of input $H\times W$.
As for the reconstruction target, we choose per-patch normalized pixels as targets with an $\normltwo$-loss, and calculate errors only on masked positions.
These designs have been proven to facilitate models to learn more informative features in \cite{mae}, and are also verified by the ablation study later in \secref{sec:app:abla}.

\paragraph{After pre-training,} we discard the decoder and only use the encoder for downstream tasks.
When fine-tuning, the pre-trained sparse encoder can be directly generalized to dense images without any tuning, due to the fact that dense input is a special case of the sparse, where every position is active.

\section{Empirical Results} \label{sec:exp}
\vspace{-1pt}
\subsection{Implementation Details} \label{sec:exp:impl}

\paragraph{Components.} SparK can use any convolutional network as the encoder, without any special design of the backbone architecture.
We implement SparK with two of the most representative convnet families: 
classical ResNets \citep{resnet} and modern ConvNeXts \citep{convnext}.
One can easily test SparK on other convolutional architectures as well.
As for the mask embeddings $\texttt{[}\texttt{M}_{4\range 1}\texttt{]}$, we implement them as random-initialized learnable feature vectors.
For decoding, we use a lightweight UNet decoder.
See \appref{sec:app:dec} for its detailed structure.
Positional embeddings are not used since convnet already encodes the spatial information. We also test this in the ablation study (\secref{sec:app:abla}).

\paragraph{Simple implementation of pre-training.}
For simplicity the same hyperparameters are used for all architectures (ResNets, ConvNeXts) and model sizes, even though tuning each may improve our fine-tuning performance at face value.
All models are pre-trained with 1.28 million unlabeled images from ImageNet-1K \citep{dataset_imagenet} training set for 1600 epochs.
Only the minimal augmentation is required (random cropping and horizontal flipping).
We use the same mask patch size (32) and ratio (60\%) as in SimMIM \citep{simmim}.
We train with a LAMB optimizer \citep{lamb}, a batch size of 4096, and a cosine-annealing learning rate with peak value $=0.0002\times batchsize/256$.

\paragraph{Fine-tuning.}
We use the official implementations of ResNet \citep{rsb}, MoCoV2 \citep{mocov2}, and ConvNeXt \citep{convnext} to fine-tune.
See \appref{sec:app:imn_ft} and \ref{sec:app:coco_ft} for recipes.

\vspace{-1pt}
\subsection{ImageNet Evaluation} \label{sec:exp:imn}

\paragraph{Performance comparison with self-supervised transformers.}
We first validate SparK on ImageNet with the pure convolutional model ConvNeXt \citep{convnext}.
Smaller models \{ViT, Swin, ConvNeXt\}-S and the bigger ones \{ViT, Swin, ConvNeXt\}-B are compared separately.
By comparing the results vertically in \tabref{tab:main}, one can find the convolutional models, with SparK pre-training, overwhelmingly outperform transformer-based pre-training methods by large margins ($+0.7\range2.7$), though SparK neither employs external models (DALL-E dVAE \citep{dalle}), nor profits from advanced (MIM+CL) pre-training.
This is somewhat surprising since transformers are well-known data-hungry models with much less inductive bias than convnets, and therefore are considered to benefit more from large-scale self-supervised training.
The result here conveys a new message:
convnets may have much more potential than expected, and their capability in visual representation may not be inferior to that of transformers.
The key may depend on how to use powerful pre-training algorithms (\eg, SparK or masked modeling) to turn this \textit{potential} into \textit{capability}.

\paragraph{Efficiency.} \label{sec:enc_eff}
Similar to MAE \citep{mae}, SparK has the advantage of encoding efficiency, especially compared to contrastive learning that encodes two or more images in a forward pass.
For instance, DINO and iBOT by default \citep{dino,ibot} use multi-crop with 2 global crops of $224\times 224$ and 10 locals of $96\times 96$, leading to $2+10\,(96/224)^2\approx3.8$ times the cost of single image encoding.
In contrast, SparK requires only 40\% of the theoretical overhead thanks to the sparsity of masked input: 60\% of patches are masked, and sparse convolution only processes the rest.
\rbt{In practice, we found a sparse ResNet-50 can save \range 23\% memory footprint (26.4 GB \textit{vs.} 34.5 GB for single batch size of 128). This allows us to train it on a 32GB Tesla V100, which otherwise is impossible for non-sparse pre-training.
The efficiency also helps SparK scale up more easily.}

\begin{table}[hb]
\vspace{-2pt}
\renewcommand\arraystretch{1.1}
\centering
{
\caption{\smallcaption
\textbf{Comparing SparK and self-supervised transformers on ImageNet.} All methods pre-train on ImageNet-1K an fine-tune with the resolution of 224.
Top-1 validation accuracy is reported, the best results are in bold.
``Extra model'' indicates whether DALL-E's dVAE (trained on 250 million extra data) is used in pre-training.
Entries with $\dag$ are quoted from \cite{ibot}.
$\ddag$ is our reproduction using the official codes.
}\label{tab:main}
\vspace{-4pt}
\scalebox{0.97}{
\begin{tabular}{l|ccc|cl|cl}
\toprule
\multirow{2}{*}{Pre-training method} &
  \multirow{2}{*}{\begin{tabular}[c]{@{}c@{}}PT\\ task\end{tabular}} &
  \multirow{2}{*}{\begin{tabular}[l]{@{}c@{}}Enc.\\ cost\end{tabular}} &
  \multirow{2}{*}{\begin{tabular}[c]{@{}c@{}}Extra~\\ model\end{tabular}} &
  \multicolumn{2}{c|}{Small backbone} &
  \multicolumn{2}{c}{Base backbone} \\
                         &        &             &      & Arch.  & Acc.          & Arch.  & Acc. \\
\tabbl
\multicolumn{8}{l}{\footnotesize{\it Vision Transformer Backbone}} \\
MoCov3~\citep{mocov3}    & CL     & 5.0$\times$ & \ncr & ViT-S  & 81.4          & ViT-B  & 83.2 \\
DINO~\citep{dino}        & CL     & 9.5$\times$ & \ncr & ViT-S  & 82.0          & ViT-B  & 82.8 \\
BEiT~\citep{beit}        & MIM    & 2.5$\times$ & \nck & ViT-S  & 81.4$^\dag$   & ViT-B  & 83.2 \\
CIM~\citep{cim}          & MIM    & 2.5$\times$ & \nck & ViT-S  & 81.6          & ViT-B  & 83.3 \\
CAE~\citep{cae}          & MIM    & 2.5$\times$ & \nck & ViT-S  & 81.8          & ViT-B  & 83.6 \\
MAE~\citep{mae}          & MIM    & 0.6$\times$ & \ncr & ViT-S  & 81.5$^\ddag$  & ViT-B  & 83.6 \\
SimMIM~\citep{simmim}    & MIM    & 2.5$\times$ & \ncr & ViT-S  & 81.7          & ViT-B  & 83.8 \\
iBOT~\citep{ibot}        & \small{MIM+CL} & 9.5$\times$  & \ncr & ViT-S  & 82.3          & ViT-B  & 84.0 \\
SimMIM~\citep{simmim}    & MIM    & 2.5$\times$ & \ncr & Swin-S & 83.4$^\ddag$  & Swin-B & 84.0 \\
\tabbl
\multicolumn{8}{l}{\footnotesize{\it Convolutional Backbone}} \\
\rowcolor{bblue!88}SparK\,(ours) & MIM    & 1$\times$   & \ncr & ConvX-S & \textbf{84.1} & ConvX-B & \textbf{84.8} \\
\bottomrule
\end{tabular}}
\vspace{-3pt}
}
\end{table}

\begin{table}[ht]
\vspace{-6pt}
\renewcommand\arraystretch{1.04}
\centering
{
\caption{\smallcaption
\textbf{Comparing convnet-based SparK with transformer-based self-supervised learning on downstream tasks.}
On ImageNet, the same fine-tuning resolution of 224 is used.
On COCO, Mask R-CNN with FPN is equally applied.
All methods follow a 3$\times$ COCO schedule (36 epochs), while MAE fine-tunes longer (50 epochs).
Average precisions of detection box (AP$^\text{bb}$) and segmentation mask (AP$^\text{mk}$) on \texttt{val2017} are reported.
$\ddag$ is reproduced using the official codes, since \cite{convnext} only runs models with Cascade Mask R-CNN.
}\label{tab:cnx_det}
\vspace{-4pt}
\setlength{\tabcolsep}{7pt}
\scalebox{0.945}{
\begin{tabular}{lcc|ccccc}
\toprule
\multirow{2}{*}{Pre-training method} &
  \multirow{2}{*}{Arch.} &
  \multirow{2}{*}{\begin{tabular}[c]{@{}c@{}}Eff.\tablefootnote{``Effective epoch'' takes into account the total amount of images processed in pre-training. For instance, a typical contrastive learning encodes two images per forward pass, so the effective epoch is twice the literal value.}\\ epoch\end{tabular}} &
  \multirow{2}{*}{\begin{tabular}[c]{@{}c@{}}Cls.\\ Acc.\end{tabular}} &
  \multicolumn{2}{c}{Det.} &
  \multicolumn{2}{c}{Seg.} \\
                                    &         &      &      & AP$^\text{bb}$ & AP$_\text{75}^\text{bb}$ & AP$^\text{mk}$ & AP$_\text{75}^\text{mk}$ \\
\midrule
MoCov3~\citep{mocov3}               & ViT-B   & 1600 & 83.2 & 47.9           & $-$                      & 42.7           & $-$                      \\
BEiT~\citep{beit}                   & ViT-B   & 800  & 83.2 & 49.8           & $-$                      & 44.4           & $-$                      \\
\midrule
Supervised~\citep{mae}              & ViT-B   & 300  & 82.3 & 47.9           & $-$                      & 42.9           & $-$                      \\
MAE~\citep{mae}                     & ViT-B   & 1600 & 83.6 & 50.3           & $-$                      & 44.9           & $-$                      \\
\textit{improvements over baseline} &         &      & \textbf{+1.3} & +2.4           & $-$                      & \textbf{+2.0}           & $-$                      \\
\midrule
Supervised~\citep{swin}             & Swin-B  & 300  & 83.5 & 48.5           & 53.2                     & 43.2           & 46.7                     \\
SimMIM~\citep{simmim}               & Swin-B  & 800  & 84.0 & 50.4           & 55.5                     & 44.4           & 47.9                     \\
\textit{improvements over baseline} &         &      & +0.5 & +1.9           & +2.3                     & +1.2           & +1.2                     \\
\midrule
Supervised$^\ddag$~\citep{convnext} & ConvX-B & 300  & 83.8 & 47.7           & 52.6                     & 43.2           & 46.6                     \\
\rowcolor{bblue!88}Spark\,(ours)                       & ConvX-B & 1600 & \textbf{84.8} & \textbf{51.2}  & \textbf{56.1}   & \textbf{45.1}  & \textbf{48.9}  \\
\textit{improvements over baseline} &         &      & +1.0 & \textbf{+3.5}           & \textbf{+3.5}                     & +1.9           & \textbf{+2.3} \\
\bottomrule
\end{tabular}}
\vspace{-8pt}
}
\end{table}

\subsection{Transferring to Downstream Tasks} \label{sec:exp:coco}
\vspace{-3pt}

Previous results on ImageNet classification have exposed the potential of SparK pre-training.
In this part we further evaluate the representation quality on fundamental downstream tasks, including object detection and instance segmentation on COCO \citep{COCO}.
These tasks are challenging, serving as professional feature evaluators because they place higher demands than classification:
models need to predict not only \textit{what}, but also \textit{where} the objects (instances) are.
Here, we consider two different settings: comparison with self-supervised vision transformers, and then with convolutional networks.
In all COCO experiments, we do not use advanced techniques such as multi-scale testing, large-scale jittering augmentation and soft-NMS.
For more details on fine-tuning, see \appref{sec:app:imn_ft} and \ref{sec:app:coco_ft}.

\paragraph{Performance \textit{vs.} self-supervised transformers.}
\Tabref{tab:cnx_det} compares the fine-tuning results on three downstream tasks: classification (Cls.), object detection (Det.), and instance segmentation (Seg.).
Among all self-supervised methods, SparK is the best performer and the only one that pre-trains a convnet.
Even when compared to the strongest SimMIM \citep{simmim} with swin-transformer, SparK still yields superior results by $+0.8$, $+0.8$, $+0.7$ on three tasks respectively.
It is particularly worth noting that without pre-training, ConvNeXt-B and Swin-B perform \textit{similarly}.
This indicates that the gains are \textit{indeed} due to our SparK pre-training rather than the backbone difference.

Overall, it can also be seen that our approach exhibits the highest improvements over supervised baselines in \tabref{tab:cnx_det} (up to $+3.5\%$).
All these observations are consistent with those in \secref{sec:exp:imn} and once again validate that the BERT-style pre-training on convolutional networks is promising.

\begin{table}[hb]
\vspace{-2pt}
\renewcommand\arraystretch{1.09}
\centering
{
\caption{\smallcaption
\textbf{ResNet-50 results on downstream tasks.}
SparK is compared to state-of-the-art contrastive learning algorithms.
For ImageNet, the same training recipe from \cite{rsb} (300-epoch fine-tuning with 224 resolution) is used.
For COCO, Mask R-CNN ResNet50-FPN is equally fine-tuned for 12 or 24 epochs (1$\times$ or 2$\times$), with average precision on \texttt{val2017} reported.
SparK is highlighted as the only \textit{generative} method.
}\label{tab:r50}
\vspace{-4pt}
\scalebox{0.955}{
\begin{tabular}{l|cc|ccccc}
\toprule
\multirow{2}{*}{Pre-training (on ResNet-50)} &
  \multirow{2}{*}{\begin{tabular}[c]{@{}c@{}}Pre-train\\ task\end{tabular}} &
  \multirow{2}{*}{\begin{tabular}[c]{@{}c@{}}Eff.\\ epoch\end{tabular}} &
  \multirow{2}{*}{\begin{tabular}[c]{@{}c@{}}Cls.\\ (Acc.)\end{tabular}} &
  \multicolumn{2}{c}{1$\times$ Schedule} &
  \multicolumn{2}{c}{2$\times$ Schedule} \\
                        &     &      &      & AP$^\text{bb}$ & AP$^\text{mk}$ & AP$^\text{bb}$ & AP$^\text{mk}$ \\
\midrule
Supervised              & $-$           & $-$  & 79.8 & 38.9       & 35.4       & 41.3   & 37.3   \\
SimSiam~\citep{simsiam} & ~Contrastive  & 800  & 79.1 & $-$        & $-$        & $-$        & $-$   \\
MoCo~\citep{moco}       & ~Contrastive  & 800  & $-$  & 38.5       & 35.1       & 40.8       & 36.9  \\
MoCov2~\citep{mocov2}   & ~Contrastive  & 1600 & 79.8 & 40.4       & 36.4       & 41.7       & 37.6  \\
SimCLR~\citep{simclr}   & ~Contrastive  & 4000 & 80.0 & $-$        & $-$        & $-$        & $-$   \\
InfoMin~\citep{infomin} & ~Contrastive  & 800  & $-$  & 40.6       & 36.7       & 42.5       & 38.4  \\
BYOL~\citep{byol}       & ~Contrastive  & 1600 & 80.0 & 40.4       & 37.2       & 42.3       & 38.3  \\
SwAV~\citep{swav}       & ~Contrastive  & 1200 & 80.1 & $-$        & $-$        & 42.3       & 38.2  \\
\rowcolor{bblue!88}SparK\,(ours) &
  Generative &
  1600 &
  \textbf{80.6} &
  \textbf{41.6} &
  \textbf{37.7} &
  \textbf{43.4} &
  \textbf{39.4}
  \\
\bottomrule
\end{tabular}}
}
\vspace{-4pt}
\end{table}

\textbf{Performance \textit{vs.}\,self-supervised convnets.}\,
We then compare SparK to state-of-the-art convolutional contrastive learning methods.
In \tabref{tab:r50}, all \textit{contrastive} methods are basically on par with supervised pre-training.
While SparK, the first \textit{generative} pre-training method for hierarchical convnets, performs significantly better than them across all downstream tasks by $+0.5\range1.2$ points.
In particular, SparK does not rely on sophisticated augmentations which have proven to be essential for contrastive learning \citep{simclr,infomin}.
We attribute these superior results to the fact that generative pre-training (SparK) can inherently provide more supervisory signals than discriminative methods: it optimizes a reconstruction loss, a form of regression loss, which is considered to be more dense and localized than contrastive learning's instance classification loss.

\paragraph{Feature transferability.}
An intriguing phenomenon in \tabref{tab:cnx_det} and \ref{tab:r50} is that the improvements over supervised baselines are more significant on COCO tasks than on ImageNet ($+3.5$ for ConvNet and $+2.7$ for ResNet).
Notice there are several key differences between these two datasets: \first the image resolution of COCO is much higher than that of ImageNet; \second most images in ImageNet are object-centric, while COCO images usually contain multiple disorganized objects.
This \textit{domain gap} poses a challenge for transfer learning, and SparK is demonstrated able to face it.
This shows SparK can learn highly transferable features through the BERT-style generative pre-training.

\vspace{-3pt}
\subsection{Scaling up SparK}  \label{sec:exp:scale}
\vspace{-3pt}

\paragraph{We gradually scale up} the model size or training resolution and test SparK's performance.
Results are reported in \tabref{tab:scale}, where we quote the accuracy of supervised baselines from \cite{rsb} (the latest ResNet baselines) and \cite{convnext} (ConvNeXt) as ``Baseline Acc.''.
As shown in the last column in \tabref{tab:scale}, one can observe that with our SparK pre-training, \textit{all} models except ResNet-50 achieve performance on par with their non-pretrained versions of larger sizes.
Such a qualitative leap indicates SparK can push a convnet to the ``next level'' in terms of representation capability.
Comparing the results horizontally, SparK improves all supervised baselines by large margins of $+0.8\range1.7$, verifying such a \textit{self-supervised} learning can make better use of model capacity than \textit{supervised} pre-training in this evaluation.
Overall, the results demonstrate a favorable scaling ability of SparK as larger models benefit more.
The steady gains across classical and modern architectures also make us believe SparK can boost many other state-of-the-art convolutional networks like VAN \citep{van}, RepLKNet \citep{replk}, and InternImage \citep{internimage}.

\renewcommand{\ncr}{\ding{55}}
\begin{table}[bh]
\renewcommand\arraystretch{1.0}
\centering
\small
{
\caption{\smallcaption
\textbf{Scaling up SparK with model size and training resolution.}
ImageNet top-1 accuracy is reported.
Absolute improvements over baselines are listed as $\Delta$.
The last column indicates whether \textit{SparK's} performance with a smaller model (\eg, 84.1 of ConvNeXt-S) reaches the \textit{baseline} of a larger one (\eg, 83.8 of ConvNeXt-B).
}\label{tab:scale}
\vspace{-4pt}
\setlength{\tabcolsep}{7pt}
\scalebox{1.0}{\begin{tabular}{l|ccc|ccc|c}
\toprule
\multicolumn{1}{l|}{Architecture} &
  \multicolumn{1}{c}{Reso.} &
  \multicolumn{1}{c}{\begin{tabular}[c]{@{}c@{}}\#Para.\\ (M)\end{tabular}} &
  \multicolumn{1}{c|}{\begin{tabular}[c]{@{}c@{}}FLOPs\\ (G)\end{tabular}} &
  \multicolumn{1}{c}{\begin{tabular}[c]{@{}c@{}}Baseline\\ Acc. \end{tabular}} &
  \multicolumn{1}{c}{\begin{tabular}[c]{@{}c@{}}\rowcolor{bblue!88}Spark\\ Acc.\end{tabular}} &
  \multicolumn{1}{c|}{\begin{tabular}[c]{@{}c@{}}$\Delta$\end{tabular}} &
  \multicolumn{1}{c}{\begin{tabular}[c]{@{}c@{}}Reach the\\next level\end{tabular}} \\
\tabbl
\multicolumn{8}{l}{\footnotesize{\emph{Classical Architecture}}}                  \\
ResNet-50       & 224 & 25.6 & 4.1  & 79.8 & 80.6 & +0.8 & \ncr   \\
ResNet-101      & 224 & 44.5 & 7.9  & 81.3 & 82.2 & +0.9 & \nck   \\
ResNet-152      & 224 & 60.2 & 11.6 & 81.8 & 82.7 & +0.9 & \nck   \\
ResNet-200      & 224 & 64.7 & 15.1 & 82.1 & 83.1 & +1.0 & $-$    \\
\tabbl
\multicolumn{8}{l}{\footnotesize{\emph{Modern Architecture}}}                \\
ConvNeXt-Small  & 224 & 50.0 & 8.7  & 83.1 & 84.1 & +1.0 & \nck \\
ConvNeXt-Base   & 224 & 89.0 & 15.4 & 83.8 & 84.8 & +1.0 & \nck \\
ConvNeXt-Large  & 224 & 198  & 34.4 & 84.3 & 85.4 & +1.1 & $-$  \\
ConvNeXt-Large  & 384 & 198  & 101  & 84.3 & \textbf{86.0} & \textbf{+1.7} & $-$ \\
\bottomrule
\end{tabular}}
\vspace{-8pt}
}
\end{table}

\subsection{Ablation Study} \label{sec:app:abla}
\vspace{-3pt}

In this study, we gradually ablate the components in SparK framework and check the corresponding performance respectively.
ImageNet fine-tuning results of each SparK's variants are listed in \tabref{tab:abla}.

\paragraph{Core designs.}
We first remove the two most important designs in SparK: sparse masking strategy and hierarchical architecture.
By replacing our sparse strategy with the zero-outing discussed in \secref{sec:method:mask}, we observe a noticeable performance degradation in \ablaref{3} of \tabref{tab:abla} that almost \textit{reaches} the supervised baseline.
This suggests the issues raised by zero-outing (like data distribution shift in \figref{fig:intro} and mask pattern vanish in \figref{fig:ssc}) can lead to ineffective pre-training.
We then remove the hierarchical design (\ablaref{4}), which results in a \textit{single-scale} masked modeling that is commonly used for transformers \citep{bert,beit,mae,simmim}.
It only uses the features at the end of encoder to reconstruct.
This modification is shown to impair the fine-tuning performance as well.
In sum, both sparse strategy and hierarchy design play key roles in SparK.

\paragraph{Other components.}
In addition, we find adding absolute positional embeddings (\ablaref{5}) is practically useless for learning convolutional representations.
We also observe calculating loss values only on masked patches gives higher accuracy (\ablaref{6}), which is consistent with \cite{mae}.
Finally and reasonably, our SparK benefits from longer pre-training as verified in \ablaref{7}.

\renewcommand{\ncr}{\ding{55}}
\begin{table}[t]
\renewcommand\arraystretch{1.0}
\centering
\small
{
\vspace{-8pt}
\caption{\smallcaption
\textbf{The ablation study on the importance of each components in SparK}.
Experiments are based on ConvNeXt-Small, with ImageNet validation accuracy reported.
Our default setting is in \ablaref{2}.
Differences are highlighted in blue.
``APE'': absolute positional embedding;
\rbt{``std.'': standard deviation of four experiments}.
}\label{tab:abla}
\vspace{-4pt}
\scalebox{0.95}{\begin{tabular}{ll|ccccc|ccc}
\toprule
            & Method               & Masking                & Hierarchy       & APE             & Loss           & Epoch          & Acc.            & $\Delta$            & \rbt{std.} \\
\midrule
\specialrule{0em}{2pt}{2pt}
\grayfg{1} & \grayfg{Not pretrained} &                      &                 &                 &                &                & \grayfg{83.1}   & \grayfg{-1.0}       &   \\
\specialrule{0em}{2pt}{1pt}
\ablanum{2} & SparK\,(ours)        & sparse                 & \nck            & \ncr            & masked only    & 1600           & 84.1            & ~0.0                & \rbt{0.07} \\
\ablanum{3} & zero-outing          & \bluebg{zero-outing}   & \nck            & \ncr            & masked only    & 1600           & 83.2            & -0.9                & \rbt{0.06} \\
\ablanum{4} & w/o hierarchy        & sparse                 & \bluebg{\footnotesize{\ncr}} & \ncr & masked only  & 1600           & 83.6            & -0.5                & \rbt{0.04} \\
\ablanum{5} & w/ APE               & sparse                 & \nck            & \bluebg{\nck}   & masked only    & 1600           & 83.9            & -0.2                & \rbt{0.10} \\
\ablanum{6} & w/ more loss         & sparse                 & \nck            & \ncr            & \bluebg{all}   & 1600           & 83.3            & -0.8                & \rbt{0.12} \\
\ablanum{7} & pre-train less       & sparse                 & \nck            & \ncr            & masked only    & \bluebg{800}   & 83.7            & -0.4                & \rbt{0.05} \\
\bottomrule
\end{tabular}}
\vspace{-6pt}
}
\end{table}

\vspace{-1pt}
\subsection{Visualization} \label{sec:app:vis}
\vspace{-1pt}

\begin{figure}[bh]
\begin{center}
   \includegraphics[width=0.89\linewidth]{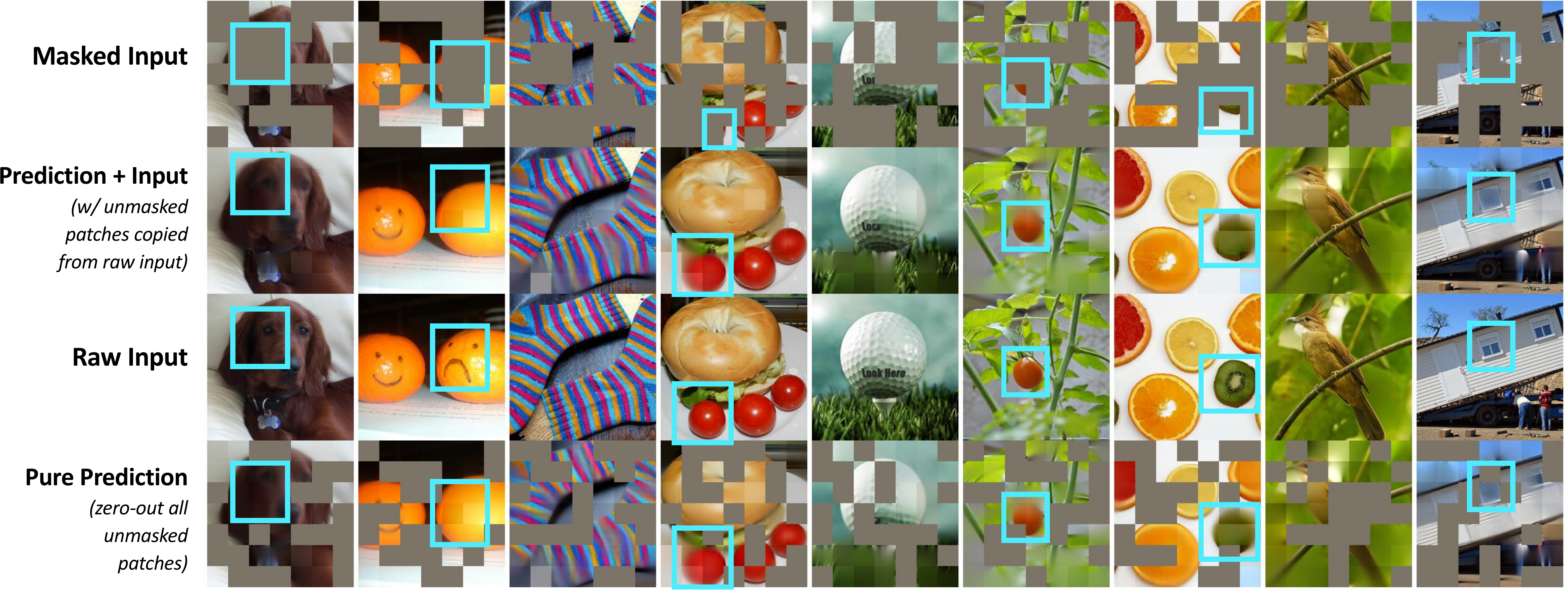}
\end{center}
\vspace{-4pt}
\caption{\smallcaption
\textbf{Reconstruction examples by a pre-trained ConvNeXt-Base with a mask ratio of $60\%$.}
Images are randomly selected from ImageNet validation set.
Several interesting regions are highlighted.
\vspace{-8pt}
}
\label{fig:vis_recon}
\end{figure}

We visualize some reconstruction results to check how the model performs in pre-training.
From \figref{fig:vis_recon} we can see that the model is able to make different but plausible predictions on masked regions (\eg, in the 2-nd column).
In the 4-th and 6-th columns, the model can almost reconstruct the round shape of red fruits from the very small portion of exposed edges.
The clear texture in the 3-rd column also shows the model can capture the visual signals with medium or high frequencies.

\vspace{-2pt}

\vspace{-4pt}
\section{Conclusion} \label{sec:conclusion}
\vspace{-2pt}
The NLP community has witnessed the rise and proliferation of masked modeling on \textit{transformers}, followed by recent efforts to validate this paradigm on \textit{vision transformers}.
However, problems arise when trying to apply this transformer-specialized algorithm to convnets.
This spurs us to take a deeper look at the fundamental differences between language and image processing, which motivates our solution.
Observing the sparse nature of point clouds coincides with masked images, we propose to treat unmasked patches as sparse voxels and use sparse convolution to encode them.
We also employ a hierarchical decoder to make full use of the advantage of convnet's hierarchy.
Our SparK makes masked modeling well suited for any convnet, and brings a performance leap on downstream tasks.
A promising future of BERT-style pre-training on convnets has been initially shown.
We hope our exploration and discovery will inspire more work to exploit the potential of convnets by generative pre-training, and thus help our vision community better embrace the pretrain-finetune paradigm.

\clearpage
{
\small
\bibliography{iclr2023_conference}

\begin{thebibliography}{64}
\providecommand{\natexlab}[1]{#1}
\providecommand{\url}[1]{\texttt{#1}}
\expandafter\ifx\csname urlstyle\endcsname\relax
  \providecommand{\doi}[1]{doi: #1}\else
  \providecommand{\doi}{doi: \begingroup \urlstyle{rm}\Url}\fi

\bibitem[Bao et~al.(2021)Bao, Dong, and Wei]{beit}
Hangbo Bao, Li~Dong, and Furu Wei.
\newblock Beit: Bert pre-training of image transformers.
\newblock \emph{arXiv preprint arXiv:2106.08254}, 2021.

\bibitem[Bateman(2014)]{bateman2014text}
John~A Bateman.
\newblock \emph{Text and image: A critical introduction to the visual/verbal
  divide}.
\newblock Routledge, 2014.

\bibitem[Bay et~al.(2006)Bay, Tuytelaars, and Gool]{surf}
Herbert Bay, Tinne Tuytelaars, and Luc~Van Gool.
\newblock Surf: Speeded up robust features.
\newblock In \emph{European conference on computer vision}, pp.\  404--417.
  Springer, 2006.

\bibitem[Brown et~al.(2020)Brown, Mann, Ryder, Subbiah, Kaplan, Dhariwal,
  Neelakantan, Shyam, Sastry, Askell, et~al.]{gpt2}
Tom Brown, Benjamin Mann, Nick Ryder, Melanie Subbiah, Jared~D Kaplan, Prafulla
  Dhariwal, Arvind Neelakantan, Pranav Shyam, Girish Sastry, Amanda Askell,
  et~al.
\newblock Language models are few-shot learners.
\newblock \emph{Advances in neural information processing systems},
  33:\penalty0 1877--1901, 2020.

\bibitem[Caron et~al.(2020)Caron, Misra, Mairal, Goyal, Bojanowski, and
  Joulin]{swav}
Mathilde Caron, Ishan Misra, Julien Mairal, Priya Goyal, Piotr Bojanowski, and
  Armand Joulin.
\newblock Unsupervised learning of visual features by contrasting cluster
  assignments.
\newblock \emph{Advances in Neural Information Processing Systems},
  33:\penalty0 9912--9924, 2020.

\bibitem[Caron et~al.(2021)Caron, Touvron, Misra, J{\'e}gou, Mairal,
  Bojanowski, and Joulin]{dino}
Mathilde Caron, Hugo Touvron, Ishan Misra, Herv{\'e} J{\'e}gou, Julien Mairal,
  Piotr Bojanowski, and Armand Joulin.
\newblock Emerging properties in self-supervised vision transformers.
\newblock In \emph{Proceedings of the IEEE/CVF International Conference on
  Computer Vision}, pp.\  9650--9660, 2021.

\bibitem[Chen et~al.(2019)Chen, Wang, Pang, Cao, Xiong, Li, Sun, Feng, Liu, Xu,
  et~al.]{mmdet}
Kai Chen, Jiaqi Wang, Jiangmiao Pang, Yuhang Cao, Yu~Xiong, Xiaoxiao Li,
  Shuyang Sun, Wansen Feng, Ziwei Liu, Jiarui Xu, et~al.
\newblock Mmdetection: Open mmlab detection toolbox and benchmark.
\newblock \emph{arXiv preprint arXiv:1906.07155}, 2019.

\bibitem[Chen et~al.(2020{\natexlab{a}})Chen, Kornblith, Norouzi, and
  Hinton]{simclr}
Ting Chen, Simon Kornblith, Mohammad Norouzi, and Geoffrey Hinton.
\newblock A simple framework for contrastive learning of visual
  representations.
\newblock In \emph{International conference on machine learning}, pp.\
  1597--1607. PMLR, 2020{\natexlab{a}}.

\bibitem[Chen et~al.(2022)Chen, Ding, Wang, Xin, Mo, Wang, Han, Luo, Zeng, and
  Wang]{cae}
Xiaokang Chen, Mingyu Ding, Xiaodi Wang, Ying Xin, Shentong Mo, Yunhao Wang,
  Shumin Han, Ping Luo, Gang Zeng, and Jingdong Wang.
\newblock Context autoencoder for self-supervised representation learning.
\newblock \emph{arXiv preprint arXiv:2202.03026}, 2022.

\bibitem[Chen \& He(2021)Chen and He]{simsiam}
Xinlei Chen and Kaiming He.
\newblock Exploring simple siamese representation learning.
\newblock In \emph{Proceedings of the IEEE/CVF Conference on Computer Vision
  and Pattern Recognition}, pp.\  15750--15758, 2021.

\bibitem[Chen et~al.(2020{\natexlab{b}})Chen, Fan, Girshick, and He]{mocov2}
Xinlei Chen, Haoqi Fan, Ross Girshick, and Kaiming He.
\newblock Improved baselines with momentum contrastive learning.
\newblock \emph{arXiv preprint arXiv:2003.04297}, 2020{\natexlab{b}}.

\bibitem[Chen et~al.(2021)Chen, Xie, and He]{mocov3}
Xinlei Chen, Saining Xie, and Kaiming He.
\newblock An empirical study of training self-supervised vision transformers.
\newblock In \emph{Proceedings of the IEEE/CVF International Conference on
  Computer Vision}, pp.\  9640--9649, 2021.

\bibitem[Choy et~al.(2019)Choy, Gwak, and Savarese]{Minkowski}
Christopher Choy, JunYoung Gwak, and Silvio Savarese.
\newblock 4d spatio-temporal convnets: Minkowski convolutional neural networks.
\newblock In \emph{Proceedings of the IEEE/CVF Conference on Computer Vision
  and Pattern Recognition}, pp.\  3075--3084, 2019.

\bibitem[Clark et~al.(2020)Clark, Luong, Le, and Manning]{electra}
Kevin Clark, Minh-Thang Luong, Quoc~V Le, and Christopher~D Manning.
\newblock Electra: Pre-training text encoders as discriminators rather than
  generators.
\newblock \emph{arXiv preprint arXiv:2003.10555}, 2020.

\bibitem[Csurka et~al.(2004)Csurka, Dance, Fan, Willamowski, and Bray]{bow}
Gabriella Csurka, Christopher Dance, Lixin Fan, Jutta Willamowski, and
  C{\'e}dric Bray.
\newblock Visual categorization with bags of keypoints.
\newblock In \emph{Workshop on statistical learning in computer vision, ECCV},
  volume~1, pp.\  1--2. Prague, 2004.

\bibitem[Dalal \& Triggs(2005)Dalal and Triggs]{HOG}
Navneet Dalal and Bill Triggs.
\newblock Histograms of oriented gradients for human detection.
\newblock In \emph{2005 IEEE computer society conference on computer vision and
  pattern recognition (CVPR'05)}, volume~1, pp.\  886--893. Ieee, 2005.

\bibitem[Deng et~al.(2009)Deng, Dong, Socher, Li, Li, and
  Fei-Fei]{dataset_imagenet}
Jia Deng, Wei Dong, Richard Socher, Li-Jia Li, Kai Li, and Li~Fei-Fei.
\newblock Imagenet: A large-scale hierarchical image database.
\newblock In \emph{2009 IEEE conference on computer vision and pattern
  recognition}, pp.\  248--255. Ieee, 2009.

\bibitem[Devlin et~al.(2018)Devlin, Chang, Lee, and Toutanova]{bert}
Jacob Devlin, Ming-Wei Chang, Kenton Lee, and Kristina Toutanova.
\newblock Bert: Pre-training of deep bidirectional transformers for language
  understanding.
\newblock \emph{arXiv preprint arXiv:1810.04805}, 2018.

\bibitem[Ding et~al.(2022)Ding, Zhang, Han, and Ding]{replk}
Xiaohan Ding, Xiangyu Zhang, Jungong Han, and Guiguang Ding.
\newblock Scaling up your kernels to 31x31: Revisiting large kernel design in
  cnns.
\newblock In \emph{Proceedings of the IEEE/CVF Conference on Computer Vision
  and Pattern Recognition}, pp.\  11963--11975, 2022.

\bibitem[Dong et~al.(2019)Dong, Yang, Wang, Wei, Liu, Wang, Gao, Zhou, and
  Hon]{unilm}
Li~Dong, Nan Yang, Wenhui Wang, Furu Wei, Xiaodong Liu, Yu~Wang, Jianfeng Gao,
  Ming Zhou, and Hsiao-Wuen Hon.
\newblock Unified language model pre-training for natural language
  understanding and generation.
\newblock \emph{Advances in Neural Information Processing Systems}, 32, 2019.

\bibitem[Dosovitskiy et~al.(2020)Dosovitskiy, Beyer, Kolesnikov, Weissenborn,
  Zhai, Unterthiner, Dehghani, Minderer, Heigold, Gelly, et~al.]{vit}
Alexey Dosovitskiy, Lucas Beyer, Alexander Kolesnikov, Dirk Weissenborn,
  Xiaohua Zhai, Thomas Unterthiner, Mostafa Dehghani, Matthias Minderer, Georg
  Heigold, Sylvain Gelly, et~al.
\newblock An image is worth 16x16 words: Transformers for image recognition at
  scale.
\newblock \emph{arXiv preprint arXiv:2010.11929}, 2020.

\bibitem[Fang et~al.(2022)Fang, Dong, Bao, Wang, and Wei]{cim}
Yuxin Fang, Li~Dong, Hangbo Bao, Xinggang Wang, and Furu Wei.
\newblock Corrupted image modeling for self-supervised visual pre-training.
\newblock \emph{arXiv preprint arXiv:2202.03382}, 2022.

\bibitem[Farabet et~al.(2010)Farabet, Martini, Akselrod, Talay, LeCun, and
  Culurciello]{farabet2010hardware}
Cl{\'e}ment Farabet, Berin Martini, Polina Akselrod, Sel{\c{c}}uk Talay, Yann
  LeCun, and Eugenio Culurciello.
\newblock Hardware accelerated convolutional neural networks for synthetic
  vision systems.
\newblock In \emph{Proceedings of 2010 IEEE International Symposium on Circuits
  and Systems}, pp.\  257--260. IEEE, 2010.

\bibitem[Felzenszwalb et~al.(2008)Felzenszwalb, McAllester, and Ramanan]{DPM}
Pedro Felzenszwalb, David McAllester, and Deva Ramanan.
\newblock A discriminatively trained, multiscale, deformable part model.
\newblock In \emph{2008 IEEE conference on computer vision and pattern
  recognition}, pp.\  1--8. Ieee, 2008.

\bibitem[Gao et~al.(2022)Gao, Ma, Li, Dai, and Qiao]{convmae}
Peng Gao, Teli Ma, Hongsheng Li, Jifeng Dai, and Yu~Qiao.
\newblock Convmae: Masked convolution meets masked autoencoders.
\newblock \emph{arXiv preprint arXiv:2205.03892}, 2022.

\bibitem[Graham \& van~der Maaten(2017)Graham and van~der Maaten]{ssc}
Benjamin Graham and Laurens van~der Maaten.
\newblock Submanifold sparse convolutional networks.
\newblock \emph{arXiv preprint arXiv:1706.01307}, 2017.

\bibitem[Grill et~al.(2020)Grill, Strub, Altch{\'e}, Tallec, Richemond,
  Buchatskaya, Doersch, Avila~Pires, Guo, Gheshlaghi~Azar, et~al.]{byol}
Jean-Bastien Grill, Florian Strub, Florent Altch{\'e}, Corentin Tallec, Pierre
  Richemond, Elena Buchatskaya, Carl Doersch, Bernardo Avila~Pires, Zhaohan
  Guo, Mohammad Gheshlaghi~Azar, et~al.
\newblock Bootstrap your own latent-a new approach to self-supervised learning.
\newblock \emph{Advances in neural information processing systems},
  33:\penalty0 21271--21284, 2020.

\bibitem[Guo et~al.(2022)Guo, Lu, Liu, Cheng, and Hu]{van}
Meng-Hao Guo, Cheng-Ze Lu, Zheng-Ning Liu, Ming-Ming Cheng, and Shi-Min Hu.
\newblock Visual attention network.
\newblock \emph{arXiv preprint arXiv:2202.09741}, 2022.

\bibitem[He et~al.(2015)He, Zhang, Ren, and Sun]{spp}
Kaiming He, Xiangyu Zhang, Shaoqing Ren, and Jian Sun.
\newblock Spatial pyramid pooling in deep convolutional networks for visual
  recognition.
\newblock \emph{IEEE transactions on pattern analysis and machine
  intelligence}, 37\penalty0 (9):\penalty0 1904--1916, 2015.

\bibitem[He et~al.(2016)He, Zhang, Ren, and Sun]{resnet}
Kaiming He, Xiangyu Zhang, Shaoqing Ren, and Jian Sun.
\newblock Deep residual learning for image recognition.
\newblock In \emph{Proceedings of the IEEE conference on computer vision and
  pattern recognition}, pp.\  770--778, 2016.

\bibitem[He et~al.(2020)He, Fan, Wu, Xie, and Girshick]{moco}
Kaiming He, Haoqi Fan, Yuxin Wu, Saining Xie, and Ross Girshick.
\newblock Momentum contrast for unsupervised visual representation learning.
\newblock In \emph{Proceedings of the IEEE/CVF conference on computer vision
  and pattern recognition}, pp.\  9729--9738, 2020.

\bibitem[He et~al.(2021)He, Chen, Xie, Li, Doll{\'a}r, and Girshick]{mae}
Kaiming He, Xinlei Chen, Saining Xie, Yanghao Li, Piotr Doll{\'a}r, and Ross
  Girshick.
\newblock Masked autoencoders are scalable vision learners.
\newblock \emph{arXiv preprint arXiv:2111.06377}, 2021.

\bibitem[Jaderberg et~al.(2015)Jaderberg, Simonyan, Zisserman, et~al.]{stn}
Max Jaderberg, Karen Simonyan, Andrew Zisserman, et~al.
\newblock Spatial transformer networks.
\newblock \emph{Advances in neural information processing systems}, 28, 2015.

\bibitem[Lin et~al.(2014)Lin, Maire, Belongie, Hays, Perona, Ramanan, Dollár,
  and Zitnick]{COCO}
Tsung-Yi Lin, Michael Maire, Serge Belongie, James Hays, Pietro Perona, Deva
  Ramanan, Piotr Dollár, and C.~Lawrence Zitnick.
\newblock Microsoft {COCO}: Common objects in context.
\newblock In \emph{ECCV}, 2014.

\bibitem[Lin et~al.(2017)Lin, Dollar, Girshick, He, Hariharan, and
  Belongie]{FPN}
Tsung-Yi Lin, Piotr Dollar, Ross Girshick, Kaiming He, Bharath Hariharan, and
  Serge Belongie.
\newblock Feature pyramid networks for object detection.
\newblock In \emph{CVPR}, 2017.

\bibitem[Liu et~al.(2015)Liu, Wang, Foroosh, Tappen, and Pensky]{spcn}
Baoyuan Liu, Min Wang, Hassan Foroosh, Marshall Tappen, and Marianna Pensky.
\newblock Sparse convolutional neural networks.
\newblock In \emph{Proceedings of the IEEE conference on computer vision and
  pattern recognition}, pp.\  806--814, 2015.

\bibitem[Liu et~al.(2016)Liu, Anguelov, Erhan, Szegedy, Reed, Fu, and
  Berg]{ssd}
Wei Liu, Dragomir Anguelov, Dumitru Erhan, Christian Szegedy, Scott Reed,
  Cheng-Yang Fu, and Alexander~C Berg.
\newblock Ssd: Single shot multibox detector.
\newblock In \emph{European conference on computer vision}, pp.\  21--37.
  Springer, 2016.

\bibitem[Liu et~al.(2019)Liu, Ott, Goyal, Du, Joshi, Chen, Levy, Lewis,
  Zettlemoyer, and Stoyanov]{roberta}
Yinhan Liu, Myle Ott, Naman Goyal, Jingfei Du, Mandar Joshi, Danqi Chen, Omer
  Levy, Mike Lewis, Luke Zettlemoyer, and Veselin Stoyanov.
\newblock Roberta: A robustly optimized bert pretraining approach.
\newblock \emph{arXiv preprint arXiv:1907.11692}, 2019.

\bibitem[Liu et~al.(2017)Liu, Cheng, Hu, Wang, and Bai]{structural2}
Yun Liu, Ming-Ming Cheng, Xiaowei Hu, Kai Wang, and Xiang Bai.
\newblock Richer convolutional features for edge detection.
\newblock In \emph{Proceedings of the IEEE conference on computer vision and
  pattern recognition}, pp.\  3000--3009, 2017.

\bibitem[Liu et~al.(2021)Liu, Lin, Cao, Hu, Wei, Zhang, Lin, and Guo]{swin}
Ze~Liu, Yutong Lin, Yue Cao, Han Hu, Yixuan Wei, Zheng Zhang, Stephen Lin, and
  Baining Guo.
\newblock Swin transformer: Hierarchical vision transformer using shifted
  windows.
\newblock In \emph{Proceedings of the IEEE/CVF International Conference on
  Computer Vision}, pp.\  10012--10022, 2021.

\bibitem[Liu et~al.(2022)Liu, Mao, Wu, Feichtenhofer, Darrell, and
  Xie]{convnext}
Zhuang Liu, Hanzi Mao, Chao-Yuan Wu, Christoph Feichtenhofer, Trevor Darrell,
  and Saining Xie.
\newblock A convnet for the 2020s.
\newblock \emph{arXiv preprint arXiv:2201.03545}, 2022.

\bibitem[Long et~al.(2015)Long, Shelhamer, and Darrell]{FCN}
Jonathan Long, Evan Shelhamer, and Trevor Darrell.
\newblock Fully convolutional networks for semantic segmentation.
\newblock In \emph{Proceedings of the IEEE conference on computer vision and
  pattern recognition}, pp.\  3431--3440, 2015.

\bibitem[Lowe(1999)]{SIFT}
David~G Lowe.
\newblock Object recognition from local scale-invariant features.
\newblock In \emph{Proceedings of the seventh IEEE international conference on
  computer vision}, volume~2, pp.\  1150--1157. Ieee, 1999.

\bibitem[Pathak et~al.(2016)Pathak, Krahenbuhl, Donahue, Darrell, and
  Efros]{inpainting1}
Deepak Pathak, Philipp Krahenbuhl, Jeff Donahue, Trevor Darrell, and Alexei~A
  Efros.
\newblock Context encoders: Feature learning by inpainting.
\newblock In \emph{Proceedings of the IEEE conference on computer vision and
  pattern recognition}, pp.\  2536--2544, 2016.

\bibitem[Radford et~al.(2019)Radford, Wu, Child, Luan, Amodei, Sutskever,
  et~al.]{gpt1}
Alec Radford, Jeffrey Wu, Rewon Child, David Luan, Dario Amodei, Ilya
  Sutskever, et~al.
\newblock Language models are unsupervised multitask learners.
\newblock \emph{OpenAI blog}, 1\penalty0 (8):\penalty0 9, 2019.

\bibitem[Ramesh et~al.(2021)Ramesh, Pavlov, Goh, Gray, Voss, Radford, Chen, and
  Sutskever]{dalle}
Aditya Ramesh, Mikhail Pavlov, Gabriel Goh, Scott Gray, Chelsea Voss, Alec
  Radford, Mark Chen, and Ilya Sutskever.
\newblock Zero-shot text-to-image generation.
\newblock In \emph{International Conference on Machine Learning}, pp.\
  8821--8831. PMLR, 2021.

\bibitem[Ronneberger et~al.(2015)Ronneberger, Fischer, and Brox]{unet}
Olaf Ronneberger, Philipp Fischer, and Thomas Brox.
\newblock U-net: Convolutional networks for biomedical image segmentation.
\newblock In \emph{International Conference on Medical image computing and
  computer-assisted intervention}, pp.\  234--241. Springer, 2015.

\bibitem[Rublee et~al.(2011)Rublee, Rabaud, Konolige, and Bradski]{orb}
Ethan Rublee, Vincent Rabaud, Kurt Konolige, and Gary Bradski.
\newblock Orb: An efficient alternative to sift or surf.
\newblock In \emph{2011 International conference on computer vision}, pp.\
  2564--2571. Ieee, 2011.

\bibitem[Sindagi et~al.(2019)Sindagi, Zhou, and Tuzel]{sp3d2}
Vishwanath~A Sindagi, Yin Zhou, and Oncel Tuzel.
\newblock Mvx-net: Multimodal voxelnet for 3d object detection.
\newblock In \emph{2019 International Conference on Robotics and Automation
  (ICRA)}, pp.\  7276--7282. IEEE, 2019.

\bibitem[Tan \& Le(2019)Tan and Le]{efficientnet}
Mingxing Tan and Quoc~V Le.
\newblock Efficientnet: Rethinking model scaling for convolutional neural
  networks.
\newblock \emph{arXiv preprint arXiv:1905.11946}, 2019.

\bibitem[Tian et~al.(2020)Tian, Sun, Poole, Krishnan, Schmid, and
  Isola]{infomin}
Yonglong Tian, Chen Sun, Ben Poole, Dilip Krishnan, Cordelia Schmid, and
  Phillip Isola.
\newblock What makes for good views for contrastive learning?
\newblock \emph{Advances in Neural Information Processing Systems},
  33:\penalty0 6827--6839, 2020.

\bibitem[Van~den Oord et~al.(2018)Van~den Oord, Li, and Vinyals]{cpc}
Aaron Van~den Oord, Yazhe Li, and Oriol Vinyals.
\newblock Representation learning with contrastive predictive coding.
\newblock \emph{arXiv e-prints}, pp.\  arXiv--1807, 2018.

\bibitem[Verelst \& Tuytelaars(2020)Verelst and Tuytelaars]{r3_cite}
Thomas Verelst and Tinne Tuytelaars.
\newblock Dynamic convolutions: Exploiting spatial sparsity for faster
  inference.
\newblock In \emph{Proceedings of the IEEE/CVF Conference on Computer Vision
  and Pattern Recognition}, pp.\  2320--2329, 2020.

\bibitem[Wang et~al.(2022)Wang, Dai, Chen, Huang, Li, Zhu, Hu, Lu, Lu, Li,
  et~al.]{internimage}
Wenhai Wang, Jifeng Dai, Zhe Chen, Zhenhang Huang, Zhiqi Li, Xizhou Zhu,
  Xiaowei Hu, Tong Lu, Lewei Lu, Hongsheng Li, et~al.
\newblock Internimage: Exploring large-scale vision foundation models with
  deformable convolutions.
\newblock \emph{arXiv preprint arXiv:2211.05778}, 2022.

\bibitem[Wei et~al.(2022)Wei, Fan, Xie, Wu, Yuille, and
  Feichtenhofer]{maskfeat}
Chen Wei, Haoqi Fan, Saining Xie, Chao-Yuan Wu, Alan Yuille, and Christoph
  Feichtenhofer.
\newblock Masked feature prediction for self-supervised visual pre-training.
\newblock In \emph{Proceedings of the IEEE/CVF Conference on Computer Vision
  and Pattern Recognition}, pp.\  14668--14678, 2022.

\bibitem[Wightman et~al.(2021)Wightman, Touvron, and J{\'e}gou]{rsb}
Ross Wightman, Hugo Touvron, and Herv{\'e} J{\'e}gou.
\newblock Resnet strikes back: An improved training procedure in timm.
\newblock \emph{arXiv preprint arXiv:2110.00476}, 2021.

\bibitem[Wu et~al.(2019)Wu, Kirillov, Massa, Lo, and Girshick]{d2}
Yuxin Wu, Alexander Kirillov, Francisco Massa, Wan-Yen Lo, and Ross Girshick.
\newblock Detectron2.
\newblock \url{https://github.com/facebookresearch/detectron2}, 2019.

\bibitem[Xie et~al.(2021)Xie, Zhang, Cao, Lin, Bao, Yao, Dai, and Hu]{simmim}
Zhenda Xie, Zheng Zhang, Yue Cao, Yutong Lin, Jianmin Bao, Zhuliang Yao,
  Qi~Dai, and Han Hu.
\newblock Simmim: A simple framework for masked image modeling.
\newblock \emph{arXiv preprint arXiv:2111.09886}, 2021.

\bibitem[Yang et~al.(2009)Yang, Yu, Gong, and Huang]{sparsecoding}
Jianchao Yang, Kai Yu, Yihong Gong, and Thomas Huang.
\newblock Linear spatial pyramid matching using sparse coding for image
  classification.
\newblock In \emph{2009 IEEE Conference on computer vision and pattern
  recognition}, pp.\  1794--1801. IEEE, 2009.

\bibitem[You et~al.(2019)You, Li, Reddi, Hseu, Kumar, Bhojanapalli, Song,
  Demmel, Keutzer, and Hsieh]{lamb}
Yang You, Jing Li, Sashank Reddi, Jonathan Hseu, Sanjiv Kumar, Srinadh
  Bhojanapalli, Xiaodan Song, James Demmel, Kurt Keutzer, and Cho-Jui Hsieh.
\newblock Large batch optimization for deep learning: Training bert in 76
  minutes.
\newblock \emph{arXiv preprint arXiv:1904.00962}, 2019.

\bibitem[Zbontar et~al.(2021)Zbontar, Jing, Misra, LeCun, and Deny]{barlow}
Jure Zbontar, Li~Jing, Ishan Misra, Yann LeCun, and St{\'e}phane Deny.
\newblock Barlow twins: Self-supervised learning via redundancy reduction.
\newblock In \emph{International Conference on Machine Learning}, pp.\
  12310--12320. PMLR, 2021.

\bibitem[Zhang et~al.(2017)Zhang, Isola, and Efros]{inpainting2}
Richard Zhang, Phillip Isola, and Alexei~A Efros.
\newblock Split-brain autoencoders: Unsupervised learning by cross-channel
  prediction.
\newblock In \emph{Proceedings of the IEEE conference on computer vision and
  pattern recognition}, pp.\  1058--1067, 2017.

\bibitem[Zhou et~al.(2021)Zhou, Wei, Wang, Shen, Xie, Yuille, and Kong]{ibot}
Jinghao Zhou, Chen Wei, Huiyu Wang, Wei Shen, Cihang Xie, Alan Yuille, and Tao
  Kong.
\newblock ibot: Image bert pre-training with online tokenizer.
\newblock \emph{arXiv preprint arXiv:2111.07832}, 2021.

\bibitem[Zhou \& Tuzel(2018)Zhou and Tuzel]{sp3d1}
Yin Zhou and Oncel Tuzel.
\newblock Voxelnet: End-to-end learning for point cloud based 3d object
  detection.
\newblock In \emph{Proceedings of the IEEE conference on computer vision and
  pattern recognition}, pp.\  4490--4499, 2018.

\end{thebibliography}
\bibliographystyle{iclr2023_conference}
}

\clearpage
\appendix
\section{Details: Decoder Architecture} \label{sec:app:dec}

SparK is a general method that does not limit the specific encoder to be pre-trained.
In other words, the definition of the encoder is all up to the user (\eg, a standard ResNet-50).
In the implementation presented in \secref{sec:exp:impl}, the only undefined component is the decoder.
We thus give its PyTorch implementation as follows.
In our experiments, the same decoder of \texttt{LightDecoder(768, 32)} is used equally for all encoders, including different ResNets and ConvNeXts.

{\footnotesize
\begin{python}
import math
import torch.nn as nn


class UNetBlock2x(nn.Module):
    def __init__(self, cin, cout):
        super().__init__()
        self.b = nn.Sequential(
            nn.Conv2d(cin, cin, kernel_size=3, stride=1, padding=1, bias=False),
            nn.BatchNorm2d(cin), nn.ReLU6(inplace=True),
            nn.Conv2d(cin, cout, kernel_size=3, stride=1, padding=1, bias=False),
            nn.BatchNorm2d(cout),
        )
    
    def forward(self, x):
        return self.b(x)


class DecoderConv(nn.Module):
    def __init__(self, cin, cout):
        super().__init__()
        self.up = nn.ConvTranspose2d(cin, cin, kernel_size=4, stride=2, padding=1, bias=True)
        
        self.conv = UNetBlock2x(cin, cout)
    
    def forward(self, x):
        x = self.up(x)
        return self.conv(x)


class LightDecoder(nn.Module):
    def __init__(self, decoder_fea_dim, upsample_ratio):
        super().__init__()
        self.fea_dim = decoder_fea_dim
        
        n = round(math.log2(upsample_ratio))
        channels = [self.fea_dim // 2 ** i for i in range(n + 1)]
        self.dec = nn.ModuleList([
            DecoderConv(cin, cout) for (cin, cout) in zip(channels[:-1], channels[1:])
        ])
        self.proj = nn.Conv2d(channels[-1], 3, kernel_size=1, stride=1, bias=True)
    
    def forward(self, to_dec):
        x = 0
        for i, d in enumerate(self.dec):
            if i < len(to_dec) and to_dec[i] is not None:
                x = x + to_dec[i]
            x = self.dec[i](x)
        return self.proj(x)

\end{python}
}

\clearpage

\section{Additional Results: Linear Evaluation}

We report the small-sized models' linear evaluation performance in \tabref{tab:le}.
In this evaluation protocol, the pre-trained backbone model is frozen and only a linear projection head would be fine-tuned.
This protocol is all the rage in contrastive learning \citep{simclr,moco,simsiam,dino}, which can probe the linear separability of deep representations.
Note that MoCoV3 \citep{mocov3} is the only contrastive learning method in \tabref{tab:le}, which aims to learn a global representation, and is therefore more suitable than non-contrastive methods on tasks like linear evaluation.
SparK shows its decent performance compared to other non-contrastive methods.

\begin{table}[ht]
\renewcommand\arraystretch{1.05}
\centering
{
\caption{\rbt{\smallcaption
\textbf{Linear evaluation results.}
Numbers of other work are directly quoted form \cite{cae}.
}}\label{tab:le}
\vspace{-2pt}
\scalebox{1.0}{
\begin{tabular}{l|ccc|c}
\toprule
Method         & BEiT &  CAE & SparK & MoCoV3 \\
\midrule
Contrastive    & \ncr & \ncr & \ncr & \nck    \\
Accuracy ($\%$) & 15.7 & 51.8 & 54.7 & 73.1   \\
\bottomrule
\end{tabular}
}
}
\end{table}

\section{Details: ImageNet Fine-tuning} \label{sec:app:imn_ft}

We refer to the latest open-source ResNet baseline of \cite{rsb} to fine-tune ResNets.
For ConvNeXts \cite{convnext}, we simply use their official implementation.
Since the original configurations in \cite{rsb,convnext} are based on supervised training from scrach, we adjust some hyperparameters for doing fine-tuning.
Details are given in \tabref{tab:hp_imn_res} and \tabref{tab:hp_imn_cnx}.

\begin{table}[ht]
\renewcommand\arraystretch{1.05}
\centering
{
\caption{\rbt{\smallcaption
\textbf{ImageNet fine-tuning recipe for ResNets, referring to \cite{rsb}.} 
}}\label{tab:hp_imn_res}
\vspace{-2pt}
\scalebox{1.0}{
\begin{tabular}{ll|ll}
\toprule
Configuration    & Value   & Configuration   & Value        \\
\midrule
Image resolution & 224     & Epochs          & 300          \\
Test image crop  & 0.95    & Batch size      & 2048         \\
Optimizer        & LAMB    & Learning rate   & 8e-3         \\
Scheduler        & Consine & Weight decay    & 0.02         \\
\midrule
Repeated aug.    & \nck  & Dropout         & \ncr           \\
Rand aug.        & 7\,/\,0.5 & Stoch. depth & \nck          \\
Gradient clip.   & \ncr  & BCE loss        & \nck           \\
Mixup alpha      & 0.1     & Label smoothing & 0.1          \\
Cutmix alpha     & 1.0     & EMA             & \{0.99,0.999\} \\
\bottomrule
\end{tabular}
}
}
\end{table}

\begin{table}[ht]
\renewcommand\arraystretch{1.05}
\centering
{
\caption{\rbt{\smallcaption
\textbf{ImageNet fine-tuning recipe for ConvNeXts, referring to \cite{convnext}.} 
}}\label{tab:hp_imn_cnx}
\vspace{-2pt}
\scalebox{1.0}{
\begin{tabular}{ll|ll}
\toprule
Configuration    & Value   & Configuration   & Value          \\
\midrule
Image resolution & 224     & Epochs          & 200            \\
Test image crop  & 0.95    & Batch size      & 2048           \\
Optimizer        & AdamW   & Learning rate   & 3.2e-3         \\
Scheduler        & Consine & Weight decay    & 0.01           \\
\midrule
Repeated aug.    & \nck  & Dropout         & \ncr             \\
Rand aug.        & 9\,/\,0.5\,/\,inc1 & Stoch. depth & \nck       \\
Gradient clip.   & \ncr  & BCE loss        & \ncr             \\
Mixup alpha      & 0.8     & Label smoothing & 0.1            \\
Cutmix alpha     & 1.0     & EMA             & \{0.99,0.999\} \\
\bottomrule
\end{tabular}
}
}
\end{table}

\section{Details: COCO Fine-tuning} \label{sec:app:coco_ft}

On COCO, we use the official implementations of MoCoV2 \citep{mocov2} and ConvNeXt \citep{convnext} to evaluate ResNets and ConvNeXts. These implementations are based on Detectron2 \citep{d2} and MMDetection \citep{mmdet} respectively.
Following the convention, we do not use advanced techniques like multi-scale testing, large-scale jittering augmentation, or soft-NMS, in all our COCO experiments for fairness. Details are in \tabref{tab:hp_coco_res} and \tabref{tab:hp_coco_cnx}.

\vspace{4pt}
\begin{table}[ht]
\renewcommand\arraystretch{1.05}
\centering
{
\caption{\rbt{\smallcaption
\textbf{COCO fine-tuning configuration for ResNets, referring to the standard implementation of MoCoV2 \citep{mocov2}.}
Mask R-CNN with FPN is used.
$x=A$ for the so-called ``A$\times$'' schedule. For instance, a 2$\times$ fine-tuning schedule means a 24-epoch training with 0.1-epoch warm-up.
}}\label{tab:hp_coco_res}
\vspace{-2pt}
\scalebox{1.0}{
\begin{tabular}{ll|ll}
\toprule
Configuration    & Value         & Configuration      & Value                   \\
\midrule
Image resize     & (384, 600)    & Normalization mean & [123.7,~116.3,~103.5]   \\
Multi-scale testing & \ncr       & Normalization std  & [58.4,~~57.1,~~57.4]    \\
Large-scale jittering aug & \ncr & Optimizer          & AdamW                   \\
Soft-NMS         & \ncr          & Weight decay       & 0.0001                  \\
Epochs           & 12$x$         & Learning rate (LR) & 2e-4                    \\
Warm-up epochs   & 0.05$x$       & LR layer decay     & 0.65                    \\
LR scheduled epochs  & [9$x$, 11$x$] & LR scheduled ratio & 0.2                 \\
\bottomrule
\end{tabular}
}
}
\end{table}

\vspace{4pt}
\begin{table}[ht]
\renewcommand\arraystretch{1.05}
\centering
{
\caption{\rbt{\smallcaption
\textbf{COCO configuration for ConvNeXts, referring to the standard implementation of ConvNeXt \citep{convnext}.}
Following the convention of self-supervised learning, 3$\times$ Mask R-CNN with FPN is used.
}}\label{tab:hp_coco_cnx}
\vspace{-2pt}
\scalebox{1.0}{
\begin{tabular}{ll|ll}
\toprule
Configuration    & Value         & Configuration      & Value                   \\
\midrule
Image resize     & (1333, 800)   & Normalization mean & [123.7,~116.3,~103.5]   \\
Multi-scale testing & \ncr       & Normalization std  & [58.4,~~57.1,~~57.4]    \\
Large-scale jittering aug & \ncr & Optimizer          & AdamW                   \\
Soft-NMS         & \ncr          & Weight decay       & 0.05                    \\
Epochs           & 36            & Learning rate (LR) & 2e-4                    \\
Warm-up epochs   & 0.15          & LR layer decay     & 0.65                    \\
LR scheduled epochs  & [27, 33] & LR scheduled ratio & 0.2                      \\
\bottomrule
\end{tabular}
}
}
\end{table}

\end{document}